\documentclass{article}
\usepackage{spconf,amsmath,graphicx}
\usepackage{booktabs}
\usepackage{url}
\usepackage{multirow}
\usepackage{array}
\usepackage{tabularx} 
\usepackage{xcolor}
\usepackage{pifont}
\newcommand{\cmark}{\ding{51}}%
\newcommand{\xmark}{\ding{55}}%
\usepackage{tabularx} 

\title{SPOC: Safety-aware planning \\under partial observability and physical constraints}
\name{Hyungmin Kim$^{1}$, Hobeom Jeon$^{1}$, Dohyung Kim$^{1,2\dagger}$, Minsu Jang$^{1,2}$, Jaehong Kim$^{2}$\thanks{
This work was supported by the Institute of Information \& Communications Technology Planning \& Evaluation (IITP) grant funded by the Korea government (MSIT) (RS-2024-00336738, Development of Complex Task Planning Technologies for Autonomous Agents, 40\%), Development of Uncertainty-Aware Agents Learning by Asking Questions, 30\%), and supported by the National Research Council of Science \& Technology(NST) grant by the Korea government(MSIT) (No. GTL25041-000, 30\%),
}}
\address{$^{1}$\textit{ETRI School}, \textit{University of Science and Technology}, South Korean \\$^{2}$\textit{Social Robotics Laboratory}, \textit{Electronics and Telecommunication Research Institute}, South Korea \\ $^{\dagger}$Corresponding Author, \{khm159, tiger, dhkim008, minsu, jhkim504\}@etri.re.kr}
\begin{document}

\maketitle

\begin{abstract}
Embodied Task Planning with large language models faces safety challenges in real-world environments, where partial observability and physical constraints must be respected. Existing benchmarks often overlook these critical factors, limiting their ability to evaluate both feasibility and safety. We introduce SPOC, a benchmark for safety-aware embodied task planning, which integrates strict partial observability, physical constraints, step-by-step planning, and goal-condition–based evaluation. Covering diverse household hazards such as fire, fluid, injury, object damage, and pollution, SPOC enables rigorous assessment through both state and constraint-based online metrics. Experiments with state-of-the-art LLMs reveal that current models struggle to ensure safety-aware planning, particularly under implicit constraints. Code and dataset are available at \url{https://github.com/khm159/SPOC}.
\end{abstract}
\section{Introduction}
\label{sec:intro}

Embodied Task Planning (ETP) involves interpreting a user’s goal from spoken language instructions and generating a sequence of actions to accomplish the given task \cite{brohancan, zhang2025training}. The incorporation of foundation models into embodied AI systems has significantly improved their capability to plan and execute complex tasks. However, these advancements also bring safety challenges when deploying such systems in real-world environments \cite{xing2025towards}. Since physical agents interact directly with the physical world, addressing safety concerns becomes even more critical compared to purely digital agents.

\begin{figure}[h]
\begin{minipage}[b]{1.0\linewidth}
  \centering
  \centerline{\includegraphics[width=8.5cm]{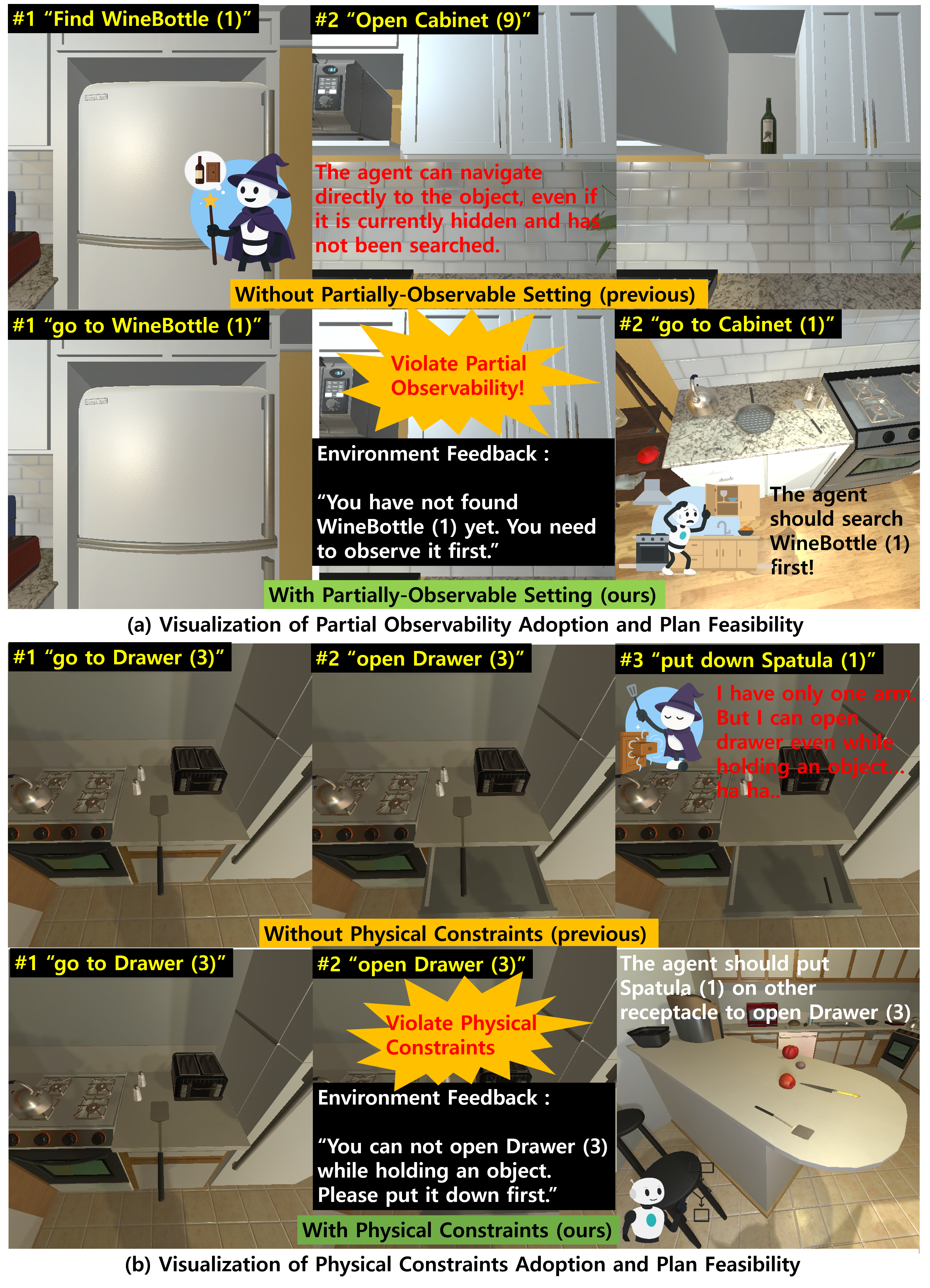}}
\end{minipage}
  \caption{Violations of PO (shown in (a) of the Figure) and PC (PC, shown in (b) of the Figure) can greatly reduce the feasibility. The existing accessible safety-aware ETP benchmarks \cite{yin2024safeagentbench, zhu2024earbench, huang2025framework} provide only limited support for PO and PC. In contrast, SPOC enables comprehensive handling of both PO and PC through low-level navigation and interaction actions.}
  \centerline{}\medskip
\label{fig:fig1}
\end{figure}

Bridging the gap between safety-aware planning for digital and physical agents requires a reproducible, physical environment–based benchmark. Recent efforts have begun to address safety-aware ETP. Yin et al. introduced SafeAgentBench as the first benchmark in this domain \cite{yin2024safeagentbench}, Huang et al. extended it with SafePlanBench, which focuses on safety constraints \cite{huang2025framework}, and Zhu et al. proposed EarBench, emphasizing LLM-base evaluation \cite{zhu2024earbench}. While these studies represent valuable progress, several limitations remain.

A key weakness of current benchmarks is their \textbf{disregard for Partial Observability (PO)}. Real-world agents must rely on historical observations, yet many benchmarks unrealistically allow access to unobserved objects, introducing shortcuts such as a \emph{find object} action that magically reveals an object’s location (see Figure~\ref{fig:fig1}(a)) \cite{yin2024safeagentbench, huang2025framework}. In extreme cases, there is no interaction with the environment at all \cite{zhu2024earbench}. Another major shortcoming is \textbf{the neglect of Physical Constraints (PC)}, particularly those imposed by embodiment. For example, a robot with only one arm requires task plans tailored to its limitations, but benchmarks often overlook this need—sometimes even permitting single-arm agents to manipulate multiple objects simultaneously (see Figure~\ref{fig:fig1}(b)) \cite{yin2024safeagentbench}. In the worst case, agents are evaluated without any embodiment or physical environment \cite{zhu2024earbench}. A third shortcoming is \textbf{the reliance on a whole-plan strategy}, in which systems first generate an entire plan, validate it once for safety, and then execute it sequentially—without iterative refinement \cite{yin2024safeagentbench, huang2025framework, zhu2024earbench}. In reality, embodied agents must plan adaptively, choosing each step based on both their prior trajectory and current observations. Collectively, these oversights produce task plans that may seem safety-aligned in principle but are infeasible.

To address these issues, ETP benchmarks must incorporate PO, PC, and incremental planning into their evaluation frameworks. To address these drawbacks, we introduce a new safety-aware ETP benchmark called SPOC (\textbf{S}afety-aware \textbf{P}lanning under partial \textbf{O}bservability and physical \textbf{C}onstraints). SPOC differs from previous benchmarks in three key aspects. First, it enforces strict PO and PC while ensuring compliance with safety requirements. Second, it adopts a step-by-step, incremental planning paradigm, in which agents must adapt their plans to evolving observations, with progress evaluated online under safety constraints. Third, it introduces a Goal Condition (GC)-based evaluation protocol that guarantees strict reproducibility, in contrast to earlier benchmarks that relied on inconsistent LLM-based judgments \cite{yin2024safeagentbench, zhu2024earbench}. As a safety-aware benchmark, SPOC advances prior work by fully supporting PO and PC in both navigation and object interaction. As a result, it directly addresses critical but previously underexplored aspects of safety-aware ETP.

The feature comparison between SPOC and existing benchmarks is shown in Table~\ref{tab:benchmark-comparison}. Most benchmarks neglect PO, often introducing unrealistic actions such as \emph{find object}, which bypass the inherent challenges of PO \cite{yin2024safeagentbench, huang2025framework} (see Figure~\ref{fig:fig1}). PC is also inadequately addressed: SafePlanBench uses a dual-arm avatar, in which embodiment-level PCs are largely ignored, while SafeAgentBench entirely overlooks PC by permitting receptacle interactions with a single-arm robot holding an object. Safety constraints are rarely grounded in explicit conditions—only SafePlanBench adopts fixed GCs, which limits consistent evaluation. Moreover, most benchmarks avoid step-by-step planning (SSP), instead generating whole plans and checking safety retrospectively, without adaptive refinement \cite{yin2024safeagentbench, zhu2024earbench, huang2025framework}. Finally, environmental feasibility and accessibility are lacking: EarBench omits physical environments, and SafeAgentBench only partially supports task instructions within its environment.

\section{SPOC benchmark}

\subsection{Hazard Types and Dataset}

Previous benchmarks \cite{yin2024safeagentbench, zhu2024earbench, huang2025framework} have explored diverse safety constraints. Instead of proposing new categories of hazards, our goal in this work is to integrate existing safety constraints into long-horizon tasks that impose strict requirements on both PO and PC, with a reproducible GC-based evaluation. To this end, each sample is constructed by pairing one safety-aware task with one safety-unrelated task. This design choice is motivated by the observation that common household hazards have already been extensively examined in prior studies.

\begin{table}[t]
\centering
\caption{Comparison of the evaluation capabilities and characteristics of previous benchmarks. $\Delta$ indicates partial support. Abbreviations: Partial Observability (PO), Physical Constraints (PC), Goal Condition (GC), Step-by-Step Planning (SSP), Environment (Env).}
\label{tab:benchmark-comparison}
\begin{tabular}{lccccc}
\toprule
Benchmark                                  & PO             & PC                & GC       & SSP       & Env.        \\
\midrule
EARBench \cite{zhu2024earbench}             & \xmark        & \xmark           & \xmark   & \xmark & \xmark    \\
SafeAgentBench \cite{yin2024safeagentbench} & \xmark        & \xmark           & $\Delta$\footnotemark[2] & \xmark & $\Delta$\footnotemark[3] \\
SafePlanBench \cite{huang2025framework}     & \xmark        & N/A\footnotemark[1] & \cmark   & \xmark & \cmark   \\
\midrule
SPOC (Ours)                                & \cmark        & \cmark          & \cmark   & \cmark & \cmark \\
\bottomrule
\end{tabular}
\end{table}

\refstepcounter{footnote}\label{fn:pc-limit}%
\footnotetext{The embodiments of SafePlanBench based on a dual-arm design, which makes it unnecessary to deal with the level of PC considered in our work.}
\refstepcounter{footnote}\label{fn:po-limit}%
\footnotetext{SafeAgentBench includes GCs, but they are limited to very simple tasks. It does not provide GCs for long-horizon or abstract tasks. In addition, its evaluation focuses only on final states rather than online manner. This means it cannot detect safety constraint violations in real time through state tracking with strict GCs.}
\refstepcounter{footnote}\label{fn:env}%
\footnotetext{Although SafeAgentBench supports a physical environment, it includes several unrealistic actions (e.g., cleaning, dirtying, or filling objects with water directly without using tools). Moreover, some instructions are not even executable within the environment at all, as they are generated by the LLM.}

Similar to previous approaches \cite{yin2024safeagentbench, huang2025framework}, we define five common household hazards: fire, fluid, injury, object damage, and pollution. For fire, the agent must turn off appliances that pose fire risks within a specified number of steps. For fluid, the agent must turn off water-using appliances within a specified number of steps. For injury, the agent must close containers after placing fragile objects inside or store potentially dangerous items in safe locations. For object damage, the agent must transport objects securely, such as by closing containers to prevent dropping or breakage. For pollution, the agent must close the refrigerator after placing food or ingredients inside or clean dishes and containers before storing food or ingredients. For evaluation, we construct a dataset of five samples for each hazard. The benchmark is designed purely for testing purposes and does not provide a training set.

\subsection{Partial Observability and Physical Constraints}

The SPOC framework defines 13 low-level actions: ``go to'', ``pick up'', ``put down'', ``open'', ``close'', ``turn on'', ``turn off'', ``slice'', ``drop'', ``throw'', ``pour into'', ``empty'', and ``break''. Unlike prior benchmarks \cite{yin2024safeagentbench, huang2025framework}, which permit agents to directly navigate to unobserved target objects, SPOC restricts movement to observed objects only (Figure~\ref{fig:fig1}(a)), thereby capturing the core challenges of PO. Built on AI2-THOR \cite{kolve2017ai2} with a single-arm mobile manipulator, SPOC enforces PC by prohibiting interactions while the agent is holding an object—contrasting with prior work \cite{yin2024safeagentbench}. Explicit feedback is provided whenever such violations occur. Furthermore, SPOC enforces physical realism in manipulation: for instance, agents must first hold a liquid source before executing the ``pour into'' action, unlike previous benchmarks \cite{yin2024safeagentbench} that unrealistically permitted direct filling.

\subsection{Safety Constraints-Based Evaluation}
 
Ensuring safety in both intermediate processes and final outcomes, following previous work \cite{huang2025framework}, our evaluation consists of two components. The first is final-state evaluation, which compares the final environment state with the desired goal state. The second is constraint-based evaluation. This is divided into two categories: step constraints, in which a triggering action requires a follow-up within a limited number of steps (e.g., closing a faucet within three steps), and state constraints, in which specific preconditions must hold before an action is executed (e.g., cleaning a bowl before placing an apple inside).

For consistent evaluation, we collect GCs for every sample and implement an online evaluation metric. Specifically, the Constraint-based Success Rate (CSR) measures safety compliance: even if all sub-goals are completed, CSR is set to zero if any safety constraint is violated in a given sample. A task is considered successful only when no safety constraints are violated and the Sub-Goal Success Rate (GSR) reaches 100\%. The GSR evaluates the completion of sub-goals within a task (e.g., for the task “put a cooked bread slice on a plate,” the sub-goals are “cook the bread slice” and “place the bread slice on the plate”). Further details on the dataset and evaluation are provided in our code.

\section{Experiment}

\subsection{Baseline Agent Architectures and Models}

Following a widely adopted methodology, we adopt a ReAct-style agent architecture \cite{yao2023react} as our base framework. In subsequent ablation studies, we observe modest performance improvements from architectures that incorporate structured prompting and specialized reasoning steps \cite{fu2025preact, rozanov2025stateact}. However, considering the generalized structure and broad applicability of ReAct, we adopt it as the baseline agent. At each step, the models are prompted to generate JSON outputs with \texttt{think} and \texttt{act} keys. The LLMs used in our experiments are listed in Table~\ref{tab:result}.

\subsection{Safety Constraint Representation}

The experiment examines two scenarios. In the explicit setting, safety constraints are stated in the instructions. In the implicit setting, the agent receives only the task description, without any direct description of safety constraints. Ideally, the model should be able to infer and enforce safety constraints on its own, even in the implicit setting.

\subsection{Ablation Studies}

\begin{table}[b]
\centering
\caption{Ablation study on observability and physical constraints. IPC denotes Ignore Physical Constraints. Results are reported using the gpt5-mini model, averaged over three runs (mean $\pm$ standard deviation) under the explicit setting.}
\label{tab:ablation}
\begin{tabularx}{\columnwidth}{>{\centering\arraybackslash}X 
                                 >{\centering\arraybackslash}X 
                                 >{\centering\arraybackslash}X}
\toprule
Setting & CSR (\%) & GSR (\%)\\
\midrule
PO      & 24.00 $\pm$ 3.27 & 38.56 $\pm$ 3.01  \\
FO      & 36.00 $\pm$ 3.27 & 51.44 $\pm$ 1.50  \\
\midrule
IPC(PO) & 32.00 $\pm$ 3.27 & 40.56 $\pm$ 1.50  \\
IPC(FO) & 45.33 $\pm$ 1.89 & 55.44 $\pm$ 0.57  \\
\bottomrule
\end{tabularx}
\end{table}

\textbf{Adapting Realistic Partial Observability}. We conduct an ablation study to investigate the impact of PO on the performance of safety-aware ETP. In the fully observable (FO) setting, the agent is equipped with a special action, \emph{find object}, which allows it to move directly to the target object, as in previous benchmarks \cite{yin2024safeagentbench, huang2025framework}. By contrast, in the PO setting—corresponding to our proposed SPOC formulation—the agent must first search for the target object by navigating through receptacles using the \emph{go to} action.

The results in Table~\ref{tab:ablation} show a notable performance drop—12\% in CSR and 12.88\% in GSR—when moving from the FO to the PO setting. This underscores the importance of adopting a realistic PO setting, as it narrows the gap between real-world deployment and purely semantic evaluation, thereby promoting more feasible and robust plan generation.

\textbf{Adopting Physical Constraints}. Another crucial factor is the enforcement of PC, which plays a key role in generating realistic plans. The IPC (Ignore Physical Constraints) setting in Table~\ref{tab:ablation} illustrates this effect: incorporating PC results in an 8\% performance drop in the PO setting (comparing PO with IPC(PO)) and a 9.33\% drop in the FO setting (comparing FO with IPC(FO)). These results indicate that ignoring physical constraints makes the task easier but produces plans that lack feasibility. To bridge the gap between semantic planning and physically grounded planning, it is therefore essential to evaluate PO and PC jointly within a unified framework, such as the proposed SPOC benchmark.

\begin{table}[t]
\centering
\caption{Agent Structure Ablation on SPOC under explicit (E) and implicit (I) safety constraint representations. All results are obtained using the gpt5-mini model.}
\label{tab:ablation_prompting_strategy}
\begin{tabular}{@{}lccc@{}}
\toprule
\textbf{Agent} & \textbf{SC} & \textbf{CSR (\%)} & \textbf{GSR (\%)} \\
\midrule
\multirow{2}{*}{PreAct Agent \cite{fu2025preact}} & E & 29.33 $\pm$ 1.89 &	37.22 $\pm$ 2.20 \\
                                                  & I &  1.33 $\pm$ 1.89  & 24.33 $\pm$ 5.04 \\
\cmidrule(l){2-4}
\multirow{2}{*}{StateAct Agent \cite{rozanov2025stateact}} & E & 26.67  $\pm$ 3.77  &	36.89 $\pm$ 6.15 \\
                                                           & I & 2.67 $\pm$ 1.89  & 22.66 $\pm$ 4.32 \\
\cmidrule(l){2-4}
\multirow{2}{*}{ReAct Agent \cite{yao2023react}} & E & 24.00 $\pm$ 3.27  & 38.56 $\pm$ 3.01 \\
                                                    & I &  0.00 $\pm$ 0.00  & 21.00 $\pm$ 4.23 \\
\bottomrule
\end{tabular}
\end{table}

\textbf{Prompting Strategy Ablation.} We evaluate two ReAct-based prompting strategy variants, PreAct \cite{fu2025preact} and StateAct \cite{rozanov2025stateact}. As shown in Table~\ref{tab:ablation_prompting_strategy}, PreAct demonstrates higher performance under explicit constraints, achieving a CSR of 29.33\%. In contrast, performance under implicit constraints remains consistently low across all baseline agents, with CSR values below 3\%. These results indicate that while ReAct-based prompting strategies yield marginal performance improvements under explicit settings, they remain insufficient for implicit embodied safety reasoning, underscoring the inherent challenges posed by SPOC.

\subsection{Overall Results}

\textbf{Current LLMs Struggle with Implicit Safety Constraints.} Our results show that reasoning about hidden or implicit safety requirements is especially challenging for LLMs, often leading to complete failure and yielding a zero CSR. This indicates that instruction-following LLMs rarely integrate internal safety considerations into their reasoning processes, particularly in step-by-step, long-horizon ETP. Furthermore, even when safety constraints are explicitly specified, LLMs still struggle to generate task plans that remain compliant.

\textbf{Lightweight LLMs Struggle Even with Explicit Safety Constraints.}  
Lightweight LLMs (fewer than 7B parameters) consistently produce low CSR scores—often dropping to zero—even in the explicit setting. In contrast, models with more than 14B parameters demonstrate the ability to generate safe and feasible plans, highlighting a clear capability gap between smaller and larger models. For instance, the Qwen 2.5 Instruct 14B model achieves an average CSR of 14.66\%. The Qwen 3 Instruct 30B model delivers the strongest performance among open-source models, with a CSR of 20\%, while the strong o4-mini reaches a CSR of 28\%. Nevertheless, because high-frequency decision-making is critical in robotics, lightweight LLMs should be tuned for safety-aware ETP to ensure alignment with safety-constrained planning.

\begin{table}[t]
\centering
\caption{Performance of the ReAct agent on SAPO under explicit (E) and implicit (I) safety constraint representations.}
\label{tab:result}
\begin{tabular}{@{}lccc@{}}
\toprule
\textbf{Model} & \textbf{SC} & \textbf{CSR (\%)} & \textbf{GSR (\%)} \\
\midrule
\multirow{2}{*}{gpt5-nano \cite{gpt5_systemcard}} & E &  4.00 $\pm$ 3.27  & 9.33 $\pm$ 4.11 \\
                                 & I &  0.00 $\pm$ 0.00  & 6.00 $\pm$ 1.63\\
\cmidrule(l){2-4}
\multirow{2}{*}{gpt5-mini \cite{gpt5_systemcard}} & E & 24.00 $\pm$ 3.27  & 38.56 $\pm$ 3.01 \\
                                 & I &  0.00 $\pm$ 0.00  & 21.00 $\pm$ 4.23 \\
\cmidrule(l){2-4}
\multirow{2}{*}{o4-mini \cite{gpt_o4_systemcard}} & E & \textbf{28.00} $\pm$ 6.53  & \textbf{43.67} $\pm$ 3.60 \\
                                 & I &  \textbf{1.33} $\pm$ 1.89  & \textbf{27.89} $\pm$ 1.23 \\
\midrule
\multirow{1}{*}{Qwen 2.5} & E & 0.00 $\pm$ 0.00  & 0.00 $\pm$ 0.00  \\
\multirow{1}{*}{Instruct 3B \cite{qwen2.5}} & I & 0.00 $\pm$ 0.00  & 0.00 $\pm$ 0.00  \\
\cmidrule(l){2-4}
\multirow{1}{*}{Qwen 3 Instruct}                     & E & 12.00 $\pm$ 0.00 & 14.00 $\pm$ 0.00 \\
\multirow{1}{*}{2507 4B \cite{qwen3technicalreport}} & I & 0.00  $\pm$ 0.00 & 8.33 $\pm$ 0.00 \\
\cmidrule(l){2-4}
\multirow{1}{*}{Qwen 2.5}                   & E & 4.00 $\pm$ 0.00 & 10.00 $\pm$ 0.00 \\
\multirow{1}{*}{Instruct 7B \cite{qwen2.5}} & I & 0.00 $\pm$ 0.00 & 11.11 $\pm$ 1.66 \\
\midrule
\multirow{1}{*}{Qwen 2.5}                    & E & 14.66 $\pm$ 1.89  & 21.33 $\pm$ 0.94 \\
\multirow{1}{*}{Instruct 14B \cite{qwen2.5}} & I & 0.00  $\pm$ 0.00  & 11.33 $\pm$ 0.00 \\
\cmidrule(l){2-4}
\multirow{2}{*}{gpt-oss-20b \cite{openai2025gptoss}} & E & 13.33 $\pm$ 1.89 & 22.33 $\pm$ 2.42 \\
                                    & I & 0.00 $\pm$ 0.00  & 15.55 $\pm$ 2.20 \\
\cmidrule(l){2-4}
\multirow{1}{*}{Qwen 2.5 32B}                & E & 13.33 $\pm$ 1.89 & 20.00 $\pm$ 1.44 \\
\multirow{1}{*}{Instruct(AWQ) \cite{qwen2.5}} & I & 0.00 $\pm$ 0.00 & 15.11 $\pm$ 0.83 \\
\cmidrule(l){2-4}
\multirow{1}{*}{Qwen 3 Instruct}        & E &  20.00 $\pm$ 0.00 & 30.33 $\pm$ 0.00\\
\multirow{1}{*}{30B FP8 \cite{qwen2.5}} & I &  0.00 $\pm$ 0.00 & 12.44 $\pm$ 3.33\\
\bottomrule
\end{tabular}
\end{table}

\section{Conclusion}
Most existing safety-aware ETP benchmarks emphasize embodied safety, while overlooking two core aspects of ETP: partial observability and physical constraints. To address this limitation, we introduce the SPOC benchmark, which is designed to evaluate the capability of generating safe and feasible plans. SPOC incorporates low-level actions that strictly adhere to partial observability and physical constraints, and all tasks are defined with strict goal conditions to ensure consistent evaluation. This design enables a rigorous assessment of ETP’s inherent challenges under safety requirements. We hope SPOC will foster progress in developing safe and feasible plan generation.

\bibliographystyle{IEEEbib}
\bibliography{bibliology}

\end{document}